\documentclass[letterpaper, 10 pt, conference]{ieeeconf}  

\IEEEoverridecommandlockouts                              

\usepackage{amsmath} 
\usepackage{amssymb}  

\usepackage{tabularx}
\usepackage{xcolor}
\newcommand{\expertcol}[1]{\textcolor{gray}{#1}}
\newcolumntype{Y}{>{\raggedleft\arraybackslash}X}
\newcolumntype{C}{>{\centering\arraybackslash}X}
\usepackage{svg}
\usepackage{booktabs}
\usepackage{subcaption}
\usepackage{multirow}
\usepackage[symbol]{parnotes}
\usepackage{siunitx}
\usepackage{nicematrix}
\usepackage{placeins}
\usepackage{tabularx}
\usepackage{array}
\usepackage{booktabs}
\usepackage{makecell}
\usepackage{censor}
\usepackage{url}
\newif\ifanonymous
\anonymousfalse 

\title{\LARGE \bf
G2DP: Diffusion Planning with Spatio-Temporal Grid Guidance
}

\ifanonymous
    \makeatletter
    \renewcommand\thanks[1]{}
    \makeatother

  \author{Anonymous Authors}
\else
    \author{%
    Hang Yu$^{1,2,\ast,\dagger}$, Ye Jin$^{3,\ast}$, Alessandro Canevaro$^{1,4}$, Julian Schmidt$^{1}$, Julian Jordan$^{1}$,\\
    Peizheng Li$^{1,4}$, Marc Kaufeld$^{3}$, Silvan Lindner$^{1}$, Johannes Betz$^{3}$, and Wilhelm Stork$^{2}$%
    \thanks{This work is a result of the research project STADT:up (19A22006O), supported by the German Federal Ministry for Economic Affairs and Energy (BMWE), based on a decision of the German Bundestag. The authors are solely responsible for the content of this publication.}%
    \thanks{$^{1}$Mercedes-Benz AG, Germany; $^{2}$Karlsruhe Institute of Technology, Germany; $^{3}$TU Munich, Germany; $^{4}$University of T{\"u}bingen, Germany}%
    \thanks{$^\ast$Equal contribution. $^{\dagger}${\tt\small hang.yu@mercedes-benz.com}}
    }
\fi

\begin{document}

\maketitle
\thispagestyle{empty}
\pagestyle{empty}

\begin{abstract}
In autonomous driving, diffusion-based planners have emerged as a promising paradigm for robust motion planning in dense and interactive traffic, as they can effectively model diverse driving behaviors.
However, their inherent stochasticity often requires explicit guidance during denoising to ensure safety and route adherence for robust closed-loop execution. Existing guidance typically relies on sparse, entity‑centric geometric queries or post-hoc refinement, yielding limited situational awareness and fragile performance in interactive scenes.
To address this issue, we propose \textit{G2DP (Grid-Guided Diffusion Planning)}, a diffusion-based planner that directly enforces dense environmental constraints through inference-time guidance. 
Specifically, G2DP constructs a differentiable spatio-temporal cost volume by fusing probabilistic future occupancy distributions with a route-progress map. 
By formulating this volume as a continuous safety energy functional, it injects dense gradients directly into the denoising loop, actively steering trajectory generation toward collision-free and progress-optimal regions.
Extensive closed-loop evaluations show that G2DP achieves state-of-the-art performance on nuPlan, outperforming the strongest imitation-learning baseline by +7.2 points in reactive score.
It further maintains top scores in zero-shot transfers to interPlan and DeepScenario benchmarks, with collision avoidance improving by +10.15 over the unguided approach on interPlan.
These results demonstrate that spatio-temporal cost grids serve as an effective representation for robust guidance in diffusion-based planning.
Code is available at: \url{https://github.com/HangYuu/G2DP}.
\end{abstract}

\section{Introduction}
\label{sec:intro}
Autonomous driving in urban traffic requires an efficient motion planner capable of modeling multi-modal human behaviors while strictly adhering to safety constraints. 
Traditional rule-based pipelines~\cite{fan2018baiduapolloemmotion, Dauner2023CORL} provide interpretability and safety guarantees, but struggle with complex interactions and long-tail scenarios, where usually extensive manual tuning is required.
Imitation-learning approaches~\cite{scheel2021urban, bansal2018chauffeurnet, vitelli2021safetynet, hallgarten2023prediction} excel in reproducing human-like driving behavior, but are sensitive to distribution shifts and may lack explicit constraint enforcement.
Recently, diffusion-based generative models~\cite{Janner2022PlanningWD, zheng2025diffusionbased} have emerged as a promising paradigm.
By reformulating planning as a conditional generation task, they capture the distribution of plausible future trajectories and, crucially, remain flexible and steerable during inference.

\begin{figure}[t]
    \centering
    \includegraphics[width=\linewidth]{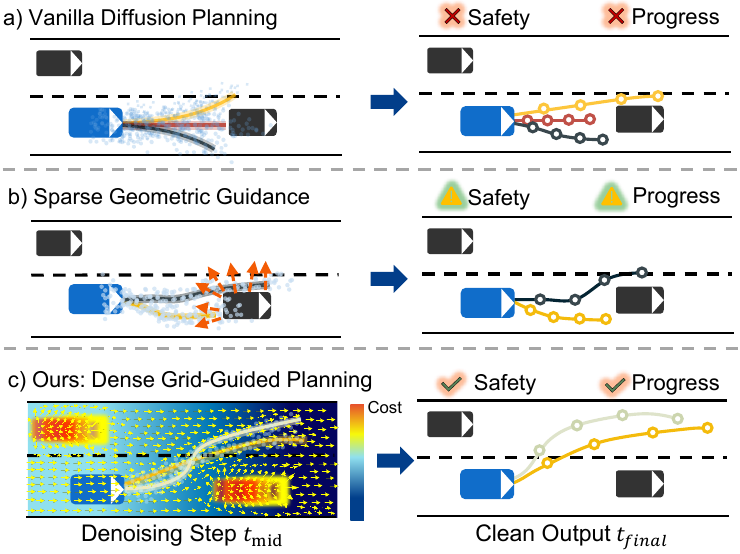}
    \caption{Diffusion guidance comparison. (a) Vanilla: Lacks explicit constraints, relying solely on learned priors. (b) Sparse: Uses instance‑anchored constraints, yielding less smooth trajectories. (c) G2DP (Ours): Employs a dense, distributional spatio‑temporal cost volume to proactively steer generation toward safe and route‑aligned energy basins.}
    \vspace{-1.5em}
    \label{fig:teaser}
\end{figure}

However, guaranteeing safety and rule compliance in closed-loop execution remains a fundamental challenge.
For diffusion-based planners, naive sampling depicted in Fig.~\ref{fig:teaser}(a) lacks explicit mechanisms to enforce safety under imperfect demonstrations.
Although diffusion models theoretically support conditional guidance, enforcing constraints effectively is non-trivial.
Current approaches typically employ sparse geometric guidance~\cite{zheng2025diffusionbased,pSTL2024,CTG2023}, such as Euclidean distances or temporal logic robustness semantics. 
These formulations evaluate constraints on a representation comprising deterministic future states for a sparse set of neighbor agents.
Their guidance is confined to instance-anchored avoidance rather than proactive steering, as illustrated in Fig. 1(b).
Alternatively, post-hoc refinement~\cite{fan2018baiduapolloemmotion,Dauner2023CORL,romer2025dpcc} reintroduces the rigidity of rule-based systems and can shift trajectories away from the learned distribution.
Consequently, incorporating dense and distributional spatial constraints directly into the differentiable generative loop remains an open problem.
Although Bird's-Eye-View (BEV) occupancy grids naturally encapsulate such spatial information, their direct formulation as differentiable cost landscapes within the denoising loop remains largely unexplored.

Inspired by this, we address this challenge by proposing \textbf{G2DP: Grid-Guided Diffusion Planning}, a planner that bridges the gap between spatio-temporal grid representations and continuous generative planning, as illustrated in Fig.~\ref{fig:teaser}(c). 
G2DP incorporates a dense and differentiable cost volume constructed by temporally aligning probabilistic future occupancy distributions and a route-progress map.
By systematically aggregating dynamic costs over the ego geometric footprint across the planning horizon, we transform this raw volume into a continuous safety energy functional. 
During denoising, we compute the analytical gradient of this energy functional with respect to the trajectory states, and inject it into the sampling update. 
This provides dense and well-conditioned gradients that proactively steer the generative process toward safe and route-aligned trajectories, obviating the need for complex post-hoc refinement.

We validate G2DP through extensive evaluations on nuPlan~\cite{nuplan}, where it achieves state-of-the-art (SOTA) performance among imitation-based planners, significantly outperforming the strongest baseline~\cite{tan2025flow} with +7.2 score improvement on the challenging Test14-hard benchmark.
We further test generalization under zero-shot transfer on interPlan~\cite{Hallgarten2024CanVM} and DeepScenario~\cite{deepscenario3d2025}, showing substantial gains in safety and route-aligned progress without tuning.

In summary, our key contributions are as follows:
\begin{itemize}
    \item We propose a novel guidance method for diffusion-based planning that overcomes the limitations of sparse, deterministic queries by injecting dense, distributional cost field directly into the denoising loop.
    
    \item We formulate a differentiable spatio-temporal energy functional that evaluates dynamic collision risks over the ego geometric footprint, enabling analytical gradients to proactively steer trajectory generation without altering the generative backbone.
    
    \item We demonstrate that G2DP achieves SOTA closed-loop performance on nuPlan among imitation-based baselines, while establishing robust zero-shot generalization capabilities in complex and interactive scenarios on interPlan and DeepScenario.
\end{itemize}

\section{Related Work}

\subsection{Optimization-Based Planning}

Classical motion planning for autonomous driving has been dominated by optimization-based approaches that enforce safety through hand-crafted constraints.
Methods such as~\cite{fan2018baiduapolloemmotion,Treiber_2000,OptimalFrenWerling} and traditional search algorithms like $A^*$~\cite{Search-Based_Optimal} provide strong interpretability and deterministic collision guarantees.
Recent extensions, such as PDM-Closed~\cite{Dauner2023CORL}, combine lane-following rules with heuristic scoring to improve benchmark performance. However, these methods rely heavily on manual cost function tuning and lack the flexibility required to handle complex multi-agent interactions.
Their predefined heuristics are difficult to scale to long-tail scenarios and tend to produce overly conservative plans in dynamic environments.

\subsection{Learning-Based Planning and Refinement}

Imitation-learning planners aim to overcome these limitations by cloning expert driving behaviors from large-scale datasets.
Early approaches employed CNN~\cite{PredictionNet, Hawke2019UrbanDW, 2019Kendalllearningtodrive} or RNN~\cite{bansal2018chauffeurnet} architectures, whereas recent work leverages Transformers to better capture global scene context and multi-agent interaction.
Representative examples include PlanTF~\cite{cheng2023plantf} and GameFormer~\cite{Huang_2023_ICCV}, which learn policies from structured environmental representations. 
Related BEV and occupancy-based models further demonstrate the effectiveness of dense spatial representations for autonomous driving scene understanding~\cite{powerbev,ago,spacedrive,liu2026driveva,wang2026linstereo,zhang2025perturbation,li2026rethinking,xiong2026unidrive}.
Despite improved expressiveness, pure behavior cloning suffers from covariate shift and does not inherently guarantee safety.
Recent analyses further show that human driving demonstrations can contain noise, outliers, and safety-rule violations, which may be inherited by imitation-based planners~\cite{trafficsafetyrulecompliance}.

Consequently, many SOTA learning-based planners operate in a learning-initialized, rule-finalized regime.
Systems such as DTPP~\cite{huang_dtpp_2024} and PLUTO~\cite{cheng2024pluto} rely on optimization-based refinement to rectify unsafe trajectories and enforce kinematic feasibility.
This dependence on hand-engineered fallback mechanisms limits the autonomy and adaptability of end-to-end learned planning pipelines.

\subsection{Diffusion-Based Planning and Guided Sampling}
Generative models, particularly diffusion probabilistic models~\cite{ho2020denoising,pmlr-v37-sohl-dickstein15}, have recently been adopted as a paradigm for motion planning since they can represent rich, multi-modal trajectory distributions through global trajectory generation.
MotionDiffuser~\cite{motiondiff} and Diffusion Planner~\cite{zheng2025diffusionbased} instantiate this idea with conditional diffusion, while flow-matching planners such as FlowPlanner~\cite{tan2025flow} follow a similar noise-to-data generation pipeline.
Compared to autoregressive planners~\cite{Sun2024GeneralizingMP,chen2024drivinggptunifyingdrivingworld}, which roll out actions sequentially and can accumulate errors over long horizons, these generative approaches improve long-horizon stability.
However, most diffusion planners~\cite{zheng2025diffusionbased,pSTL2024,CTG2023} operate primarily on vectorized data, restricting their guidance to sparse geometric constraints rather than dense environmental perception.

\subsection{Guidance Mechanisms}
Controllability during inference is a key advantage of diffusion models. Existing guidance strategies in planning can be broadly categorized into sparse and dense approaches.

\noindent\textbf{Sparse Geometric Guidance.} 
The majority of diffusion-based planners~\cite{zheng2025diffusionbased, pSTL2024, CTG2023} employ classifier or energy-based guidance defined on vectorized geometric primitives. 
For example, Diffusion Planner~\cite{zheng2025diffusionbased} proposes constructing energy potentials based on Euclidean distances to lane polylines or polygon boundaries. 
In the robotics domain, methods like CoBL-Diffusion~\cite{MizutaLeung2024} and SafeFlow~\cite{dai2025safeflow} integrate Control Barrier Functions to enforce safety.
Although differentiable, these sparse deterministic primitives filter out spatial uncertainty, failing to capture future distributions.
Furthermore, prior work~\cite{zheng2025diffusionbased} demonstrates guidance efficacy qualitatively in isolated scenarios, lacking systematic closed-loop quantitative ablation on large-scale benchmarks.
Other methods like classifier-free guidance~\cite{ho2021classifierfree} provide coarse mode steering but cannot enforce fine-grained safety constraints.

\begin{figure*}[htbp]
    \centering
    \includegraphics[width=\textwidth]{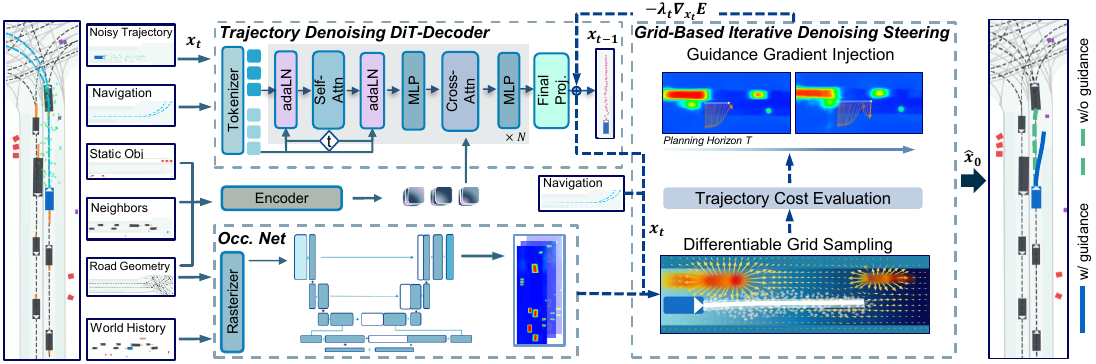}
    \caption{Model architecture of the G2DP. 
    The system utilizes a DiT-Decoder to process ego future noisy trajectories conditioned on vectorized scene context, alongside an Occupancy Network that predicts occupancy distributions from rasterized history.
    Crucially, our Grid-Based Iterative Denoising Steering (right) bridges these components. 
    It queries the predicted cost grids at the coordinates of $\mathbf{x}_t$ via differentiable sampling to compute a safety energy term. 
    A guidance gradient $-\lambda_t \nabla_{\mathbf{x}_t} E$ is then injected into the sampling step, actively steering the trajectory away from high-cost regions and preserving route progress.}
    \vspace{-1.0em}
    \label{fig:model_architecture}
\end{figure*}

\noindent\textbf{Dense Grid-Based Guidance.} 
In contrast, grid representations can parameterize the environment as a dense probabilistic field.
Hybrid methods such as HYPE~\cite{yu2025hypehybrid} and TPP~\cite{Chen2023TreestructuredPP} incorporate occupancy grids but apply them via discrete tree search refinement, which lacks differentiability.
In the domain of visual navigation, NaviDiffusor~\cite{NaviDiffusor} utilizes depth-derived cost maps for guidance. 
However, it relies on a single local and static depth projection.
Our approach distinguishes itself by leveraging spatio-temporal BEV occupancy probabilities to construct a time-varying cost volume. 
This allows us to inject time-aligned, ego-geometry-aware cost gradients directly into the denoising loop, enabling proactive safety in dynamic and interactive traffic scenes.

\section{Methodology}
We present G2DP, a grid-guided diffusion planner that unifies trajectory generation with explicit safety constraints through a differentiable spatio-temporal cost volume. As depicted in Fig.~\ref{fig:model_architecture}, our approach integrates a Diffusion Transformer (DiT)~\cite{dit_2023_ICCV} backbone to capture the distribution of ego future trajectories. In parallel, a U-Net--based predictor~\cite{ronneberger_u-net_2015} estimates future occupancy probability grids, which are subsequently fused with spatial maps encoding forward progress along the navigation route. This fusion forms a dense, differentiable guidance field encoding both collision likelihood and navigation desirability.
This parallel design allows the diffusion model to specialize in trajectory synthesis, as the grid predictor focuses on temporally aggregated spatial risk estimation.
At inference time, the predicted cost grids establish a global and spatially continuous energy field, yielding dense guidance gradients at denoising steps to shape the sampling process toward trajectories that jointly minimize risk and maintain progress.
The following subsections detail the diffusion-based trajectory generator, the cost grid construction, and the grid-guided sampling procedure, respectively.
In the paper, $t$ denotes the diffusion step, and $\tau$ denotes the temporal index of the future trajectory.

\subsection{Diffusion-Based Trajectory Generation}
\label{sec:Diffusion_Based}
\noindent\textbf{Trajectory Representation.}
We formulate motion planning as conditional trajectory generation for the ego vehicle over a fixed horizon of $T_{\text{fut}}$ steps.
The ego state at each timestep is represented by its 2D position and heading, parameterized as
$(x_\tau, y_\tau, \cos\theta_\tau, \sin\theta_\tau)$.
We represent the ego trajectory as a sequence of states $\mathbf{x} \in \mathbb{R}^{(T_{\text{fut}}+1)\times 4}$ by stacking the current state and $T_{\text{fut}}$ future states, and treat $\mathbf{x}$ as the primary generation target for the diffusion model.

\noindent\textbf{Scene Encoding.}
Following recent diffusion-based planners~\cite{zheng2025diffusionbased}, we encode the surrounding scene into a compact set of tokens.
Past neighbor states, static objects, and lane segments are processed by dedicated fusion encoders,
yielding a set of scene context tokens $\mathcal{S}$ that summarize surrounding agents and map geometry.

\noindent\textbf{DiT-Based Diffusion Model.}
We adopt a DiT~\cite{dit_2023_ICCV} backbone that operates directly in the trajectory space.
We model a variance-preserving diffusion process over $\mathbf{x}$, where noise is gradually added in the forward process, and a neural network learns to denoise in the reverse process.
At each diffusion step $t$, the noisy trajectory $\mathbf{x}_t$ is projected into a latent embedding and enriched with an agent-type token that distinguishes the ego from context entities.
We embed the diffusion step $t$ and combine it with the encoded navigation route.
The core denoising network is a stack of DiT blocks with adaptive layer normalization and attention modulation.
Each block applies self-attention to the ego noisy trajectory tokens and cross-attention to the scene context.

\noindent\textbf{Sampling and Ego-Only Generation.}
During training, we sample steps $t$, obtain the corresponding noisy trajectory $\mathbf{x}_t$ from the forward process, and supervise the DiT backbone to predict the denoising target.
At inference time, we initialize the process by concatenating the current ego state with Gaussian noise for the future steps, and iteratively solve the reverse diffusion using a DPM-Solver~\cite{Ludpmsolver2022}.
A strict constraint corrector enforces that the first state in $\mathbf{x}_t$ always matches the current ego state. Past neighbor trajectories are used only as conditioning signals in the context encoder and are not generated by the model.
The output is a single ego trajectory proposal $\hat{\mathbf{x}}_0$ consistent with the learned scene distribution.

\subsection{Cost Grid Map}
\label{sec:Occupancy Probablity Map}

We employ a lightweight U-Net-based model~\cite{ronneberger_u-net_2015} to predict spatio-temporal occupancy grids from observed history and static scene context. These grids are then converted into a differentiable cost volume that provides dense guidance for the diffusion sampler in Sec.~\ref{sec:Diffusion_Based}, and can be queried by arbitrary trajectory samples during denoising.

\noindent\textbf{Input Representation.}
All inputs follow an ego-centric BEV rasterization.
The dynamic history is represented as
$\mathbf{P} \in \mathbb{R}^{C_{\text{dyn}} \times T_{\text{hist}} \times H \times W}$, storing the past $T_{\text{hist}}$ agent footprints.
while the static map tensor $\mathbf{M} \in \mathbb{R}^{C_{\text{stat}} \times H \times W}$ encodes lane geometry, directions, drivable areas, and lane markings.

\noindent\textbf{Occupancy Prediction.}
Given the BEV inputs $(\mathbf{P}, \mathbf{M})$, the U-Net outputs future occupancy grids
$\mathbf{O}\in\mathbb{R}^{T_{\text{occ}}\times H\times W}$, 
with $O_\tau(u,v)\in[0,1]$ indicating the occupancy probability at future timestep $\tau$.
The resulting $\mathbf{O}$ serves as a dense cost field, differentiable w.r.t. trajectory states,
allowing continuous space queries to provide guiding gradients.

\noindent\textbf{Route-Aligned Progress.}
We rasterize the navigation route polyline into the ego-centric grids and compute a distance transform to the route centerline.
We then define a route desirability map $R(u,v)$ that assigns lower cost to cells that are close to the route and lie ahead along the route direction, encouraging forward progress while staying route-aligned.

\noindent\textbf{Fused Cost Grids.}
For each future timestep $\tau$, we fuse occupancy and progress into a single cost grid
\begin{equation}
\psi_\tau(u,v) = \gamma \cdot O_\tau(u,v) + (1-\gamma) \cdot R(u,v),
\label{eq:fused-cost-grid}
\end{equation}
where the occupancy weight $\gamma\in[0,1]$
balances collision avoidance and route progress.
We denote the resulting cost-grid stack as $\mathcal{C}=\{\psi_\tau\}_{\tau=1}^{T_{\text{occ}}}$, which is queried by trajectory samples during subsequent guided denoising.

\subsection{Trajectory Generation Guided by Cost Grids}
\label{sec:traj_guidance}
To enable the diffusion sampler to actively avoid obstacles while maintaining progress, we require a guidance signal that is both spatially continuous and differentiable. We leverage the fused cost grids $\mathcal{C}=\{\psi_\tau\}_{\tau=1}^{T_{\text{occ}}}$ from Sec.~\ref{sec:Occupancy Probablity Map} to define a dense energy functional. This energy steers the sampler (Sec.~\ref{sec:Diffusion_Based}) toward collision-free, progress-seeking trajectories over the aligned horizon $T=T_{\text{occ}}=T_{\text{fut}}$.

\noindent\textbf{Spatio-Temporal Energy Formulation.}
We construct an energy field that captures the vehicle's spatial geometry and temporal evolution. Evaluating safety based on a single center-point ignores the vehicle's dimensions, whereas taking the maximum cost over the footprint is sensitive to localized prediction artifacts. To balance safety coverage with robustness, we propose a Top-K footprint aggregation strategy.

Let $\mathbf{x}=\{\mathbf{x}_\tau\}_{\tau=0}^{T}$ be an ego future trajectory sample. For each timestep $\tau$, we determine the oriented rectangular footprint $\mathcal{K}_{\theta_\tau}$ based on the pose $(x_\tau, y_\tau, \theta_\tau)$ and bilinearly sample the fused cost grid $\psi_\tau$ within this region to obtain a set of costs $\mathcal{V}_\tau$. We define the scalar cost $c^{(\tau)}$ as the mean of the $K$ highest values in $\mathcal{V}_\tau$:
\begin{equation}
c^{(\tau)}
=
\frac{1}{K}\sum_{v\in\Omega_K} v,
\label{eq:footprint-cost}
\end{equation}
where $\Omega_K \subset \mathcal{V}_\tau$ represents the subset of the $K$ largest cost values. This strategy ensures guidance focuses on critical hazard boundaries without being dominated by outliers or single-pixel noise.

To ensure the planner anticipates future risks instead of responding myopically, we aggregate these per-timestep costs into a cumulative trajectory cost, which serves as the potential energy $E(\mathbf{x})$ for the diffusion process:
\begin{equation}
E(\mathbf{x})
=
\sum_{\tau=1}^{T} c^{(\tau)}.
\label{eq:trajectory-energy}
\end{equation}
The gradient $\nabla_{\mathbf{x}}E(\mathbf{x})$ thus provides a holistic spatio-temporal signal that enables global trajectory optimization.

\noindent\textbf{Gradient-Based Denoising Steering.}
During inference-time sampling, we employ classifier guidance~\cite{dhariwal2021diffusion} to enforce these constraints without retraining the backbone. Given the current noisy trajectory $\mathbf{x}_t$, we treat the original marginal distribution as a prior and reweight it with our safety energy to define a target distribution $p_t^\star$:
\begin{equation}
    p_t^\star(\mathbf{x}_t \mid \mathcal{S}, \mathcal{C})
    \propto
    p_t(\mathbf{x}_t \mid \mathcal{S})\,
    \exp\!\bigl(-\lambda_t E(\mathbf{x}_t;\mathcal{C})\bigr).
    \label{eq:guided-target-ours}
\end{equation}
Here, $p_t(\mathbf{x}_t \mid \mathcal{S})$ denotes the unconstrained marginal distribution learned by DiT.
$p_t^\star(\mathbf{x}_t \mid \mathcal{S}, \mathcal{C})$ represents the refined posterior distribution conditioned on the dense cost volume $\mathcal{C}$ (Sec.~\ref{sec:Occupancy Probablity Map}), which encodes the safety constraints.

By taking the score (gradient of log-likelihood) of Eq.~\eqref{eq:guided-target-ours}, we obtain the guided sampling update:
\begin{equation}
    \nabla_{\mathbf{x}_t} \log p_t^\star
    =
    \nabla_{\mathbf{x}_t} \log p_t
    -
    \lambda_t \nabla_{\mathbf{x}_t} E(\mathbf{x}_t;\mathcal{C}),
    \label{eq:guided-score-ours}
\end{equation}
where $\lambda_t$ is a guidance scale controlling the correction strength. The term $-\lambda_t \nabla_{\mathbf{x}_t} E$ acts as an energy-based correction injected into the DPM-Solver. This mechanism effectively steers the generation process toward the low-energy basins of $p_t^\star$, ensuring trajectories satisfy safety requirements and preserving the kinematic constraints captured by $p_t$.

This energy correction is applied during the late stages of the denoising process. Since early diffusion steps establish the macroscopic trajectory topology, injecting gradients too early can disrupt the model's behavioral intent. Restricting guidance to the final stages allows us to leverage the dense cost volume for fine-grained steering.

\subsection{Implementation Details}
\label{sec:impl_details}
\noindent\textbf{Training Strategy.}
The occupancy U-Net and diffusion model are trained separately on the nuPlan~\cite{nuplan} 1M training split, following~\cite{zheng2025diffusionbased}, and coupled only at inference. The U-Net is trained offline with focal loss supervised by future rasterized occupancy labels and remains frozen. The DiT backbone is trained via MSE loss for ego trajectory generation.

\noindent\textbf{Planning and Sampling.}
We plan for a horizon of $T_{\text{fut}}=3\,\text{s}$ at $10\,\text{Hz}$ and run a 10-step DPM-Solver at inference.

\noindent\textbf{Occupancy Prediction.}
The occupancy network predicts $30$ future occupancy grids, aligned with $10\,\text{Hz}$ on a $400{\times}400$ BEV raster at $0.25\,\text{m/px}$ ($100\,\text{m}\times100\,\text{m}$). Its dynamic input uses $T_{\text{hist}}=2\,\text{s}$ plus static map channels.

\noindent\textbf{Cost Fusion.}
We fix the occupancy weight to $\gamma=0.95$ in Eq.~\eqref{eq:fused-cost-grid} across all benchmarks.
The sensitivity of this parameter, which directly trades off collision risk against route progress, is validated in Fig.~\ref{fig:occ_share}.

\noindent\textbf{Top-$K$ Aggregation.}
For the footprint cost in Eq.~\eqref{eq:footprint-cost}, we set $K{=}15$, representing the top $\sim\!8\%$ of $\sim\!190$ footprint cells at $0.25\,\text{m/px}$. As validated in Table~\ref{tab:ablation_topk_wide}, this ensures sufficient risk coverage while mitigating sensitivity to single-cell noise.

\noindent\textbf{Guidance Scale and Injection Window.}
We restrict grid-based guidance to the final denoising stages (steps 8--9 out of 10) during inference. We apply a fixed scale of $\lambda_t=0.5$, an optimal configuration validated in Fig.~\ref{fig:guid_scale}.

\noindent\textbf{Inputs.}
G2DP uses $2\,\text{s}$ neighbor histories, with current-state-only ego input. We also report G2DP$^\dagger$, which removes all histories and uses only current ego and neighbor states.

\section{Experiments}

\subsection{Benchmarks}
We evaluate our approach on three challenging closed-loop planning benchmarks.
All experiments use the nuPlan simulator with an LQR controller in both non-reactive (NR) and reactive (R) modes.

\noindent\textbf{nuPlan.}
We evaluate our approach on nuPlan~\cite{nuplan} using three closed-loop benchmarks covering 14 scenario types:
Val14~\cite{Dauner2023CORL} (1.1k scenarios), Test14-random~\cite{cheng2023plantf} (challenge split), and Test14-hard~\cite{cheng2023plantf} (curated difficult scenarios).

\noindent\textbf{interPlan.} 
To assess robustness, we evaluate the nuPlan-trained model in a zero-shot setting on interPlan~\cite{Hallgarten2024CanVM}, without any training or hyperparameter tuning.
This benchmark comprises 335 handcrafted, highly interactive and adversarial scenarios designed to stress-test safety.

\noindent\textbf{DeepScenario.} 
To evaluate robustness against distribution shifts in real-world dense traffic, we utilize DeepScenario~\cite{deepscenario3d2025}, a drone-recorded urban dataset. We convert logs to the nuPlan format and evaluate on 411 feasible scenarios after filtering out invalid initialization states, e.g., initial collisions or off-road starts. All results are reported in a zero-shot setting using the model previously trained on nuPlan.

\subsection{Baseline Models}
We compare our method against rule-based, hybrid and learning-based methods.

\noindent\textbf{Rule-Based.}
We include the classical IDM~\cite{Treiber_2000} and the nuPlan challenge winner PDM~\cite{Dauner2023CORL}.

\noindent\textbf{Learning-Based.}
We compare against PlanTF~\cite{cheng2023plantf}, PLUTO~\cite{cheng2024pluto}, GameFormer~\cite{Huang_2023_ICCV}, Flow Planner~\cite{tan2025flow}, and Diffusion Planner~\cite{zheng2025diffusionbased}.
These baselines cover diverse imitation learning paradigms, from direct trajectory regression to generative models.
Diffusion Planner shares a similar base generative pipeline, enabling a direct assessment of the gains from our grid guidance.

\subsection{Results and Analysis}
\noindent\textbf{nuPlan Main Results.}
Table~\ref{tab:g2dp_results} summarizes closed-loop results on Val14, Test14-hard, and Test14 under NR and R settings.
G2DP achieves the strongest performance among all purely imitation-based planners (no post-hoc refinement), consistently outperforming baselines such as Diffusion Planner and Flow Planner. 
Notably, on the challenging reactive Test14-hard benchmark, G2DP${}^\dagger$ delivers significant gains, improving the reactive driving score by +8.4 compared to Diffusion Planner.
G2DP${}^\dagger$ removes all history inputs and uses only current states (Sec.~\ref{sec:impl_details}), yielding stronger reactive performance (+3.7 over G2DP) by eliminating temporal correlations.
For fair comparison with hybrid methods, we integrate the nuPlan-tuned refinement module (+ref.)~\cite{Dauner2023CORL,sun2024statetran2}. Although peak scores on the reactive Test14-hard split converge after post-hoc refinement, G2DP${}^\dagger$ exhibits a substantially reduced dependency (+4.1) compared to Diffusion Planner (+12.8), suggesting that our grid-guided sampling aligns with rule-based safety criteria.

\begin{table}[t]
\centering
\caption{Closed-loop performance on nuPlan in non-reactive (NR) and reactive (R) settings (best in bold).}
\label{tab:g2dp_results}
\setlength{\tabcolsep}{4pt}
{\scriptsize
\resizebox{\columnwidth}{!}{%
\begin{tabular}{llcccccc}
\toprule
\multirow{2}{*}{Type} & \multirow{2}{*}{Planner} & \multicolumn{2}{c}{Val14} & \multicolumn{2}{c}{Test14-hard} & \multicolumn{2}{c}{Test14} \\
\cmidrule(lr){3-4}\cmidrule(lr){5-6}\cmidrule(lr){7-8}
 &  & NR & R & NR & R & NR & R \\
\midrule

\expertcol{\makecell[l]{Expert}} & \expertcol{Log-replay}
& \expertcol{93.53} & \expertcol{80.32}
& \expertcol{85.96} & \expertcol{68.80}
& \expertcol{94.03} & \expertcol{75.86} \\
\midrule

\multirow{7}{*}{\makecell[l]{Rule-\\based\\\&\\Hybrid}}
& IDM & 75.60 & 77.33 & 56.15 & 62.26 & 70.39 & 74.42 \\
& PDM-Closed & 92.84 & 92.12 & 65.08 & 75.19 & 90.05 & 91.63 \\
& PDM-Hybrid & 92.77 & 92.11 & 65.99 & 76.07 & 90.10 & 91.28 \\
& GameFormer & 79.94 & 79.78 & 68.70 & 67.05 & 83.88 & 82.05 \\
& PLUTO & 92.88 & 92.06 & 80.08 & 76.88 & 92.23 & 90.29 \\
& Diffusion Planner +ref. & 94.26 & 92.90 & 78.87 & \textbf{82.00} & \textbf{94.80} & 91.75 \\
& Flow Planner +ref. & \textbf{94.31} & 92.38 & 78.64 & 80.25 & 94.79 & 92.40 \\
\cmidrule(lr){2-8}
& G2DP${}^\dagger$ +ref. (Ours) & 94.04 & \textbf{92.92} & \textbf{80.76} & 81.74 & 94.67 & \textbf{92.63}  \\
\midrule

\multirow{8}{*}{\makecell[l]{Learning-\\based}}
& PDM-Open & 53.53 & 54.24 & 33.51 & 35.83 & 52.81 & 57.23 \\
& GameFormer* & 13.32 & 8.69 & 7.08 & 6.69 & 11.36 & 9.31 \\
& PlanTF & 84.27 & 76.95 & 69.70 & 61.61 & 85.62 & 79.58 \\
& PLUTO* & 88.89 & 78.11 & 70.03 & 59.74 & 89.90 & 78.62 \\
& Diffusion Planner & 89.87 & 82.80 & 75.99 & 69.22 & 89.19 & 82.93 \\
& Flow Planner & 90.43 & 83.31 & 76.47 & 70.42 & 89.88 & 82.93 \\
\cmidrule(lr){2-8}

& G2DP (Ours) & \textbf{92.26} & 83.91 & \textbf{80.05} & 73.91 & \textbf{91.58} & 85.26 \\
& G2DP$^\dagger$ (Ours) & 91.08 & \textbf{86.67} & 78.77 & \textbf{77.61} & 88.54 & \textbf{88.42} \\
\bottomrule
\end{tabular}%
}}
\begin{flushleft}
\vspace{-0.5em}
\footnotesize
* w/o post-hoc refine. +ref.: w/ post-hoc refine.\\
$^\dagger$ Neighbor-agent history stacks are removed from the model input.\\

\vspace{-1.5em}
\end{flushleft}
\end{table}

\begin{table}[tb]
\centering
\caption{Performance of different planners on interPlan.}
\label{tab:interplan_results}
\setlength{\tabcolsep}{4pt}
{\small
\begin{tabularx}{\columnwidth}{lCCCC}
\toprule
\textbf{Planner} & \textbf{Overall} & \textbf{Nudge Around} & \textbf{High Traffic} & \textbf{Jaywalk} \\
\midrule
PlanTF & 47.70 & 49.40 & 58.85 & 33.94 \\
PLUTO (w/o refine.) & 58.47 & 71.56 & 67.25 & 25.48 \\
Diffusion Planner & 52.90 & 60.48 & 49.71 & 26.20 \\
Flow Planner & 61.82 & \textbf{72.96} & 67.21 & 43.57 \\
G2DP (Ours) & 61.74 & 66.71 & 67.30 & \textbf{51.12} \\
G2DP${}^\dagger$ (Ours) & \textbf{62.55} & 72.47 & \textbf{68.25} & 45.27 \\
\bottomrule
\end{tabularx}
}
\vspace{-1.3em}
\end{table}

\begin{table}[tb]
\centering
\caption{InterPlan evaluation via nuPlan metrics. Values in parentheses indicate the gap relative to Diffusion Planner.}
\label{tab:grid_guidance_benchmark}
\setlength{\tabcolsep}{2pt}
\renewcommand{\arraystretch}{1.0}
{\scriptsize
\begin{tabularx}{\columnwidth}{l CCC}
\toprule
\textbf{Metric} & \textbf{Diffusion Planner} & \textbf{G2DP} & \textbf{G2DP${}^\dagger$} \\
\midrule
Score & 52.90 & 61.74~{\color{teal}\scriptsize{($+$8.84)}} & \textbf{62.55}~{\color{teal}\scriptsize{($+$9.65)}} \\
Drivable area & 89.85 & 91.94~{\color{teal}\scriptsize{($+$2.09)}} & \textbf{93.73}~{\color{teal}\scriptsize{($+$3.88)}} \\
Driving direction & 99.10 & 97.91~{\color{red}\scriptsize{($-$1.19)}} & \textbf{99.25}~{\color{teal}\scriptsize{($+$0.15)}} \\
Comfort & 53.43 & 53.10~{\color{red}\scriptsize{($-$0.33)}} & \textbf{58.21}~{\color{teal}\scriptsize{($+$4.78)}} \\
Progress & \textbf{94.63} & 93.73~{\color{red}\scriptsize{($-$0.90)}} & 94.43~{\color{red}\scriptsize{($-$0.20)}} \\
Along expert route & 55.17 & \textbf{55.96}~{\color{teal}\scriptsize{($+$0.79)}} & 55.21~{\color{teal}\scriptsize{($+$0.04)}} \\
Collision avoidance & 84.78 & 92.84~{\color{teal}\scriptsize{($+$8.06)}} & \textbf{94.93}~{\color{teal}\scriptsize{($+$10.15)}} \\
Speed limit & \textbf{98.52} & 98.35~{\color{red}\scriptsize{($-$0.17)}} & 98.21~{\color{red}\scriptsize{($-$0.31)}} \\
TTC & 83.58 & 93.13~{\color{teal}\scriptsize{($+$9.55)}} & \textbf{94.33}~{\color{teal}\scriptsize{($+$10.75)}} \\
\bottomrule
\end{tabularx}
}
\vspace{-1.2em}
\end{table}

\begin{figure*}[t]
    \centering
    \includegraphics[width=\textwidth]{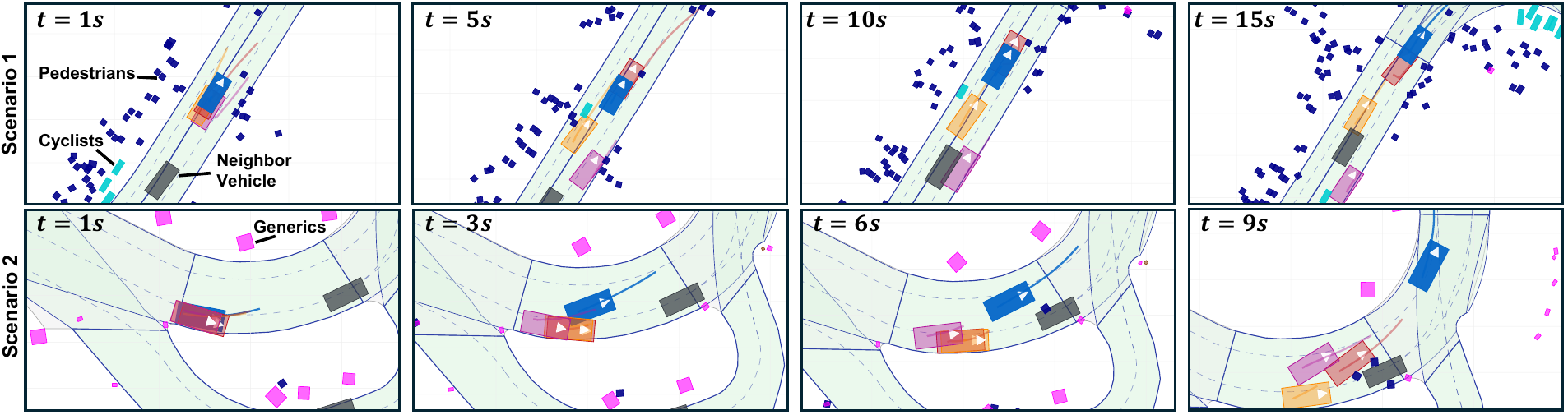}
    \vspace{-1.5em}
    \caption{Qualitative closed-loop comparisons of \textcolor[HTML]{0466C8}{\textbf{G2DP (ours)}}, \textcolor[HTML]{FF8C00}{\textbf{PlanTF}}, \textcolor[HTML]{b5179e}{\textbf{PLUTO*}}, and \textcolor[HTML]{c1121f}{\textbf{Diffusion Planner}} with planned trajectories. 
    \textbf{Top (DeepScenario):} Only G2DP maintains efficient forward progress without collisions in highly interactive traffic. 
    \textbf{Bottom (nuPlan):} Only G2DP makes a timely lane change and safely overtakes the stopped vehicle.}
    \vspace{-1.0em}
    \label{fig:results}
\end{figure*}

\noindent\textbf{interPlan Results.}
Table~\ref{tab:interplan_results} shows that G2DP outperforms Diffusion Planner by a substantial margin (+8.84). The history-free variant, G2DP${}^\dagger$, further elevates the overall score to 62.55 (+9.65), dominating the highly interactive subsets. 
The metric breakdown in Table~\ref{tab:grid_guidance_benchmark} indicates that these gains stem from larger safety margins, with the most notable improvement in Collision avoidance (+10.15) and TTC (+10.75), consistent with dense and probabilistic cost-grid guidance in highly interactive scenes.

\begin{figure}[b]
    \vspace{-0.5em}
    \centering
    \includegraphics[width=\linewidth]{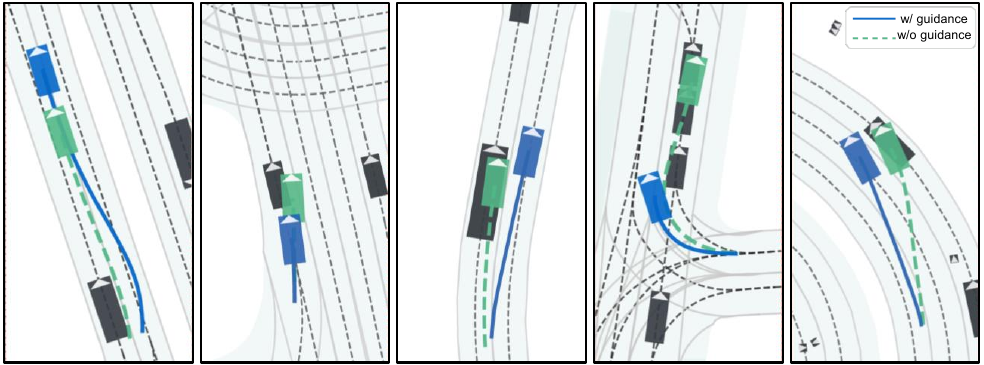}
    \vspace{-1.5em}
    \caption{Comparison of closed-loop execution with and without grid guidance.
    The vehicle boxes indicate final ego poses, and trailing curves show the executed closed-loop trajectory.
    Solid blue: with guidance. Dashed green: without guidance.}
    \label{fig:guidance_compare}
\end{figure}

\noindent\textbf{DeepScenario Results (NR).}
Table~\ref{tab:deepscenario_metrics_placeholder_nr} reports zero-shot closed-loop evaluation on DeepScenario (NR), where rule annotations are limited and ego properties vary across scenarios.
We mainly focus on safety and progress-related metrics instead of overall score and comfort.
Compared to Diffusion Planner, G2DP improves both safety and efficiency, with about +11.0 in collision avoidance and +10.5 in progress, and it also exceeds PDM-Closed on progress and route-aligned progress in the high-density urban subset.

\newcommand{\xx}{\texttt{xx.xx}}

\begin{table}[tb]
\centering
\caption{DeepScenario closed-loop NR evaluation. Focus on safety and progress-related metrics.}
\label{tab:deepscenario_metrics_placeholder_nr}
\setlength{\tabcolsep}{2pt}
{\scriptsize
\begin{tabularx}{\columnwidth}{l*{6}{C}}
\toprule
\textbf{Metric} &
\expertcol{\textbf{Expert-log}} &
\textbf{PDM-Closed} &
\textbf{PlanTF} &
\textbf{PLUTO*} &
\textbf{Diffusion Planner} &
\textbf{G2DP (Ours)}\\
\midrule
Drivable area       & \expertcol{100.00} & 69.39  & 62.34 & \textbf{70.31} & 49.24  & 68.48 \\
Driving direction   & \expertcol{96.71}  & \textbf{92.21} & 86.63 & 88.60 & 87.55  & 89.54 \\
Progress            & \expertcol{100.00} & 68.90 & 48.81 & 43.97 & 64.42  & \textbf{74.88} \\
Along expert route  & \expertcol{99.44}  & 46.00 & 31.51 & 37.70 & 47.52  & \textbf{54.36} \\
Collision avoidance & \expertcol{100.00} & 80.04 & 73.85 & 73.51 & 70.34  & \textbf{81.33} \\
TTC                 & \expertcol{79.56}  & 64.74 & 57.06 & 58.64 & 60.05  & \textbf{66.72} \\
\bottomrule
\end{tabularx}
}
\vspace{-1.3em}
\end{table}

\noindent\textbf{Qualitative Analysis.}
Fig.~\ref{fig:results} compares non-reactive closed-loop rollouts of G2DP with learning-based baselines, illustrating how G2DP maintains safety and forward progress in urban interactions. In the crowded street (Scenario 1), baselines suffer from inefficient navigation or inevitable collisions, while G2DP maintains safe and smooth progress among dense pedestrians. In Scenario 2, only G2DP executes an anticipatory lane change to safely bypass a stopped vehicle. These examples highlight that our grid-guided sampling provides spatio-temporal risk awareness, steering trajectories through complex topologies.

\begin{figure}[b]
    \vspace{-0.5em}
    \centering
    \includegraphics[width=\linewidth]{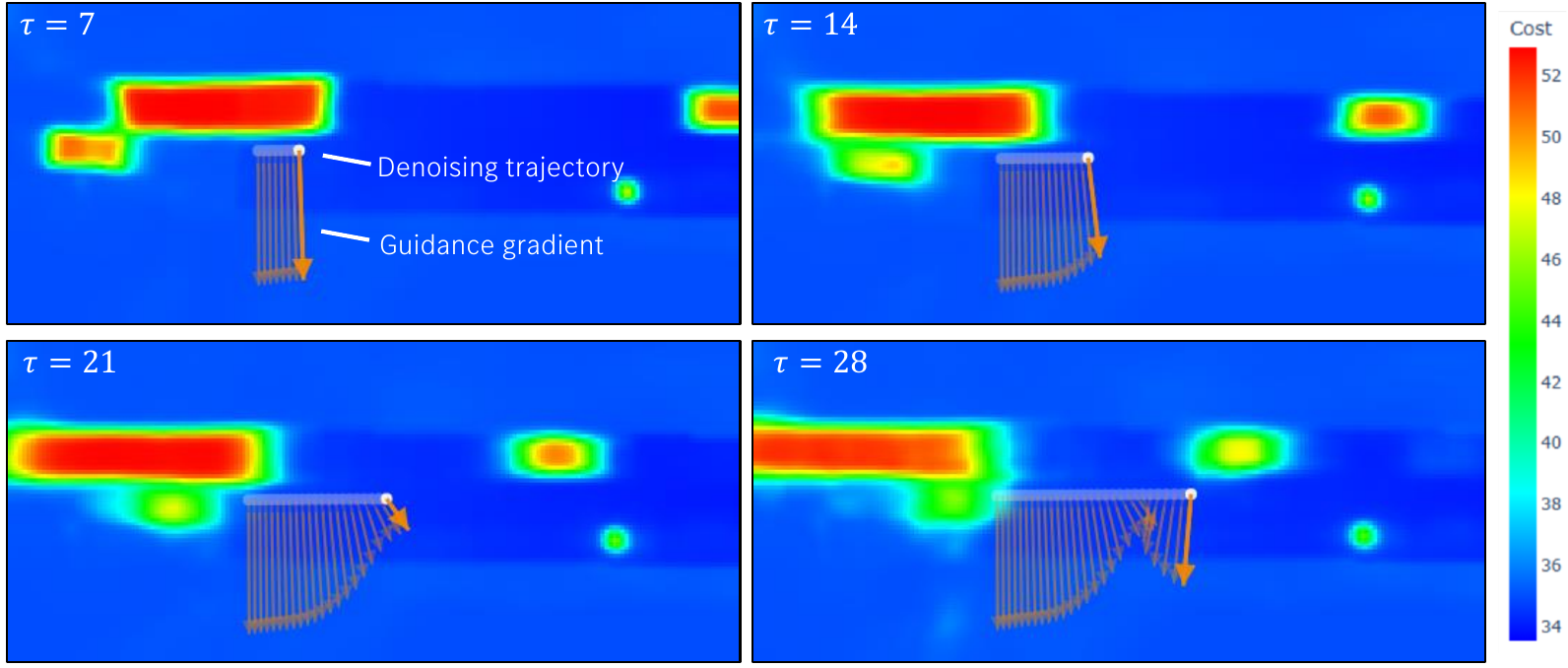}
    \vspace{-1.0em}
    \caption{Cost grid guidance at denoising step $t{=}9$.
    Each panel shows the BEV cost grid $\psi_\tau$ and the guidance gradients evaluated at a selected trajectory timestep $\tau\in\{7,14,21,28\}$.
    White dots denote the current denoising trajectory $\mathbf{x}_t$, and the arrows visualize the corresponding guidance gradients that push the trajectory toward lower cost regions.}
    \label{fig:guidance_force}
\end{figure}

Fig.~\ref{fig:guidance_compare} presents an ablation of grid guidance by comparing trajectories from the same diffusion planning backbone under identical initial conditions.
Without guidance, the planner tends to generate trajectories that drift into regions with high collision likelihood, especially in dense interactive scenarios.
In contrast, when guided by the cost grid map, the generated trajectories preserve kinematic feasibility while steering away from obstacles and conflict zones.
Remarkably, guided trajectories do not exhibit overly conservative behavior. Instead, they maintain forward progress and adhere to the intended navigation lanes.
This qualitative comparison highlights that the designed grid guidance not only suppresses collisions but reshapes the diffusion sampling process toward trajectories that jointly balance safety and progress.

Fig.~\ref{fig:guidance_force} illustrates the temporal evolution of the guidance gradients derived from the cost map during the denoising process.
Since the occupancy network outputs a sequence of future occupancy probabilities $\{\psi_\tau\}_{\tau=1}^{T}$, the resulting cost field forms a time-varying cost volume over the planning horizon.
Accordingly, the grid guidance evaluates each trajectory point against the cost map slice at its corresponding timestamp, yielding a temporally aligned guidance signal.
High-risk regions generate repulsive gradients, while low-cost areas aligned with the route provide attraction.
As a result, trajectory points are pushed away from high-risk areas and redistributed toward safer and more goal-consistent regions.
This behavior demonstrates that the grid guidance operates as a dense and continuous correction field, enabling smooth and anticipatory trajectory steering during denoising.

\subsection{Ablation Studies}
\label{sec:ablation}
\noindent\textbf{Effect of Grid Guidance.}
Table~\ref{tab:grid_guidance_metrics} ablates guidance modalities.
Grid guidance achieves the highest Score ($+3.35$) and boosts safety (Collision $+3.59$).
Conversely, sparse guidance, implemented via instance-level distance functions as in~\cite{zheng2025diffusionbased}, degrades the overall Score ($-1.96$) and Comfort ($-10.13$), even when utilizing oracle neighbor trajectories from log replay. 
Its sharp comfort drop and marginal collision gain ($+0.64$) indicate that instance-level signals induce late-stage abrupt maneuvers.
This suggests that cost grids, by encoding dense probabilistic risk distributions, enable smooth and anticipatory collision avoidance.

\begin{table}[tb]
\centering
\caption{Metrics on Test14-hard (NR). We compare G2DP using None, Sparse, and Grid guidance. Values in parentheses denote the performance change relative to the None.}
\label{tab:grid_guidance_metrics}
\setlength{\tabcolsep}{2pt}
\renewcommand{\arraystretch}{1.0}
{\scriptsize
\begin{tabularx}{\columnwidth}{l CCC}
\toprule
\textbf{Metric} & \textbf{None} & \textbf{Sparse (Oracle)} & \textbf{Grid (Ours)} \\
\midrule
Score & 76.70 & 74.74 {\color{red}\scriptsize{($-$1.96)}} & \textbf{80.05} {\color{teal}\scriptsize{($+$3.35)}} \\
Drivable area & 93.75 & 93.48 {\color{red}\scriptsize{($-$0.27)}} & \textbf{94.85} {\color{teal}\scriptsize{($+$1.10)}} \\
Driving direction & 98.34 & 97.79 {\color{red}\scriptsize{($-$0.55)}} & \textbf{98.90} {\color{teal}\scriptsize{($+$0.56)}} \\
Comfort & \textbf{93.38} & 83.25 {\color{red}\scriptsize{($-$10.13)}} & 91.92 {\color{red}\scriptsize{($-$1.46)}} \\
Progress & 97.52 & 97.79 {\color{teal}\scriptsize{($+$0.27)}} & \textbf{98.90} {\color{teal}\scriptsize{($+$1.38)}} \\
Along expert route & 87.90 & 87.84 {\color{red}\scriptsize{($-$0.06)}} & \textbf{90.18} {\color{teal}\scriptsize{($+$2.28)}} \\
Collision avoidance & 89.24 & 89.88 {\color{teal}\scriptsize{($+$0.64)}} & \textbf{92.83} {\color{teal}\scriptsize{($+$3.59)}} \\
Speed limit & 96.07 & \textbf{96.12} {\color{teal}\scriptsize{($+$0.05)}} & 96.00 {\color{red}\scriptsize{($-$0.07)}} \\
TTC & 80.88 & 81.14 {\color{teal}\scriptsize{($+$0.26)}} & \textbf{85.82} {\color{teal}\scriptsize{($+$4.94)}} \\
\bottomrule
\end{tabularx}
}
\vspace{-1.5em}
\end{table}

\begin{figure}[b]
    \centering
    \includegraphics[width=\linewidth]{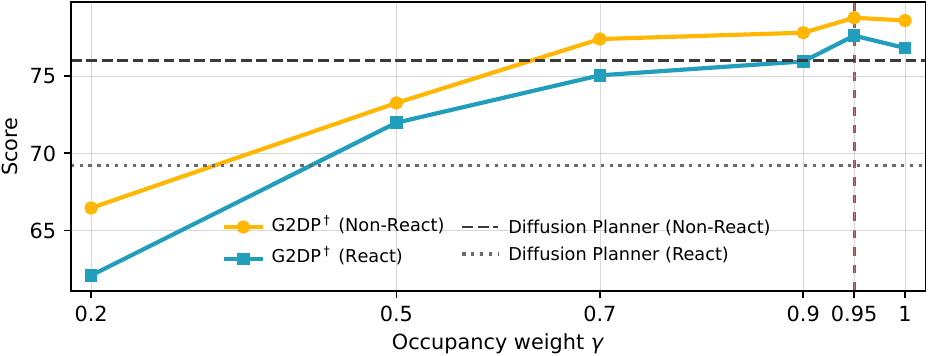}
    \caption{Impact of the occupancy weight on the Test14-hard.}
    \label{fig:occ_share}
\end{figure}

\noindent\textbf{Occupancy Weight.}
Fig.~\ref{fig:occ_share} studies the occupancy weight $\gamma$ in Eq.~\eqref{eq:fused-cost-grid}. Performance peaks at $\gamma=0.95$. Lower values, e.g., $\gamma=0.2$ inevitably weaken safety enforcement. Removing the progress term ($\gamma=1.0$) induces overly conservative braking, which increases rear-end collisions in high-density regions under non-reactive mode.

\begin{figure}[t]
    \centering
    \includegraphics[width=\linewidth]{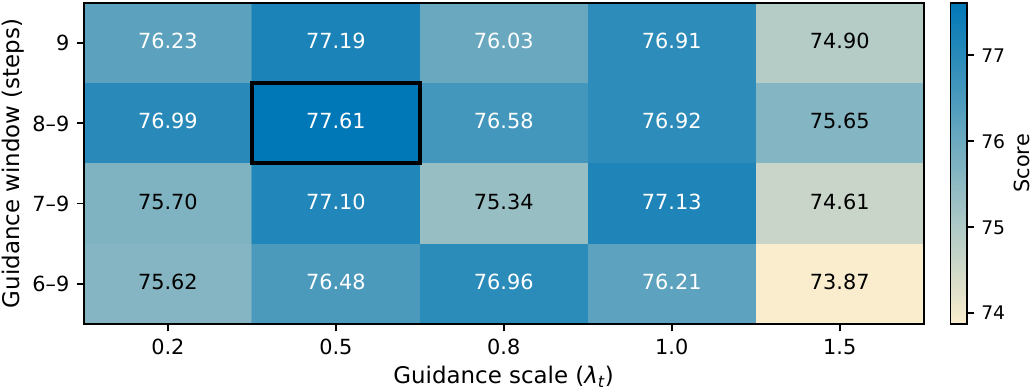}
    \vspace{-1.7em}
    \caption{Guidance scale and window ablation on Test14-hard (R). The heatmap reports reactive scores for combinations of the guidance scale $\lambda_t$ and the guidance window.}
    \vspace{-1.3em}
    \label{fig:guid_scale}
\end{figure}

\noindent\textbf{Guidance Scale and Injection Window.}
To validate the late-stage injection strategy introduced in Sec.~\ref{sec:traj_guidance}, we jointly ablate the guidance scale $\lambda_t$ and the injection window. Fig.~\ref{fig:guid_scale} reports Test14-hard (R) scores over the 2D sweep.

Overall, applying guidance closer to the final stage is consistently more effective than extending it to earlier steps.
The best performance is achieved with the Steps~8--9 window and $\lambda_t{=}0.5$ (77.61).
A smaller scale $\lambda_t{=}0.2$ provides weaker gains, suggesting that the guidance signal is insufficient to reliably correct unsafe samples.
In contrast, larger scales $\lambda_t{\ge}0.8$ tend to degrade performance across windows, indicating that overly strong gradients can over-constrain sampling and harm closed-loop behavior.

Similarly, extending the window to earlier steps, e.g., Steps~6--9 does not improve reactive performance and can be detrimental, confirming our methodology that early guidance interferes with trajectory stabilization.
This restricted Steps~8--9 setting provides robust performance gains with minimal computational overhead.

\begin{table}[tb]
  \centering
  \caption{Effect of footprint cost aggregation strategies.}
  \label{tab:ablation_topk_wide}
  \resizebox{\linewidth}{!}{%
  \begin{tabular}{l cccccc}
    \toprule
    \textbf{Split (Score $\uparrow$)} & Average & Max & Top-5 & Top-10 & Top-15 & Top-20 \\
    \midrule
    Val14 (NR)      & 86.33 & 90.58 & 90.71
 & 91.25 & \textbf{92.26} & 91.68 \\
    Test14-hard (NR)& 77.07 & 76.84 & 77.73 & 79.88 & \textbf{80.05} & 78.71 \\
    \bottomrule
  \end{tabular}%
  }
\end{table}

\noindent\textbf{Cost Aggregation.} We investigate footprint cost aggregation strategies in Table~\ref{tab:ablation_topk_wide}. Average pooling (86.33 on Val14) lets safe cells wash out collision signals. Conversely, max pooling (76.84 on Test14-hard) is sensitive to localized prediction artifacts. Our Top-$K$ strategy in Eq.~\eqref{eq:footprint-cost} balances this trade-off, peaking at $K=15$, which captures hazard boundaries and filters single-pixel noise.

\section{Conclusion}
We presented G2DP, a grid-guided diffusion planner that effectively bridges the gap between dense environmental context and continuous trajectory generation. By transforming time-varying occupancy probabilities into a differentiable spatio-temporal cost volume, our approach injects dense safety gradients that consistently steer the generative sampling process toward low-cost basins. Extensive closed-loop evaluations on the nuPlan benchmark, complemented by zero-shot transfers to the interactive interPlan and DeepScenario datasets, demonstrate that G2DP establishes a new SOTA among imitation-learning planners. The results consistently show that dense grid-based guidance significantly enhances safety-critical metrics and reactive robustness in complex urban topologies. 
This work validates that spatio-temporal, probabilistic cost volumes can robustly enforce strict safety in generative motion planning.

\noindent\textbf{Limitations and Future Work.}
G2DP relies on a simplified vehicle footprint and requires tuning for guidance scheduling. Future work will explore an end-to-end framework, directly mapping raw sensors to cost volumes to unify perception and guided generative planning within a single loop.

\addtolength{\textheight}{-1.3cm}   


\bibliographystyle{IEEEtran}
\bibliography{IEEEabrv, HIUVT, references_zotero, related_work}

\end{document}